\documentclass[sigconf, nonacm]{acmart}
\renewcommand\footnotetextcopyrightpermission[1]{}
\usepackage{multirow}
\usepackage{arydshln}
\usepackage{hyperref}
\usepackage{pifont}
\usepackage{cleveref}

\AtBeginDocument{%
  }

\copyrightyear{2025}
\acmYear{2025}
\begin{document}

\title{Multi-Task Dense Prediction Fine-Tuning with Mixture of Fine-Grained Experts}

\author{Yangyang Xu}
\affiliation{%
  \institution{Department of Computer Science and Technology, Tsinghua University}
  \state{Beijing}
  \country{China}
}
\email{yangyangxu@mail.tsinghua.edu.cn}

\author{Xi Ye}
\authornote{Xi Ye is the corresponding author.}
\affiliation{%
  \institution{Department of Computer Science and Technology, Tsinghua University}
  \state{Beijing}
  \country{China}}
\email{xi_ye@mail.tsinghua.edu.cn}

\author{Duo Su}
\affiliation{%
  \institution{Department of Computer Science and Technology, Tsinghua University}
  \state{Beijing}
  \country{China}}
\email{suduo@mail.tsinghua.edu.cn}

\renewcommand{\shortauthors}{Yangyang Xu et al.}

\begin{abstract}
Multi-task learning (MTL) for dense prediction has shown promising results but still faces challenges in balancing shared representations with task-specific specialization.
In this paper, we introduce a novel Fine-Grained Mixture of Experts (FGMoE) architecture that explores MoE-based MTL models through a combination of three key innovations and fine-tuning.
First, we propose intra-task experts that partition along intermediate hidden dimensions of MLPs, enabling finer decomposition of task information while maintaining parameter efficiency. Second, we introduce shared experts that consolidate common information across different contexts of the same task, reducing redundancy, and allowing routing experts to focus on unique aspects. Third, we design a global expert that facilitates adaptive knowledge transfer across tasks based on both input feature and task requirements, promoting beneficial information sharing while preventing harmful interference.
In addition, we use the fine-tuning approach to improve parameter efficiency only by training the parameters of the decoder.
Extensive experimental results show that the proposed FGMoE uses fewer parameters and significantly outperforms current MoE-based competitive MTL models on two dense prediction datasets (\textit{i.e.,} NYUD-v2, PASCAL-Context) in various metrics.
\end{abstract}



\maketitle

\section{Introduction}\label{sec:intro}

Multi-task learning (MTL) refers to the simultaneous learning of multiple related dense prediction tasks, enabling a model to jointly estimate various outputs, where each pixel or region in an image needs to be classified or predicted in a context-sensitive manner, such as semantic segmentation, depth estimation, and boundary detection\cite{MTL_survey_2021}. 
Classic MTL methods\cite{kokkinos2017cnn_MTL} learn a shared representation via CNN/Transformer-based model among different tasks and use task-specific heads.
This approach not only promotes shared representations among tasks, leading to more efficient learning, but also enhances the model's ability to perform holistic vision-based scene understanding.

Learning task-specific features is a fundamental problem for MTL models. To achieve this objective, recent MTL methods\cite{atrc_2021,InvPT_2022,xu2023demt,taskexpert23} design a decoder-focused paradigm that exchanges information during the decoding stage. Most decoder-focused methods use a shared backbone and design different decoders for each task.
However, the decoder-focused MTL methods for dense prediction often suffer from inefficiency in handling task-specific complexities, particularly when learning multi-scale features and structured multi-scale task interactions.
The decoder-focused approaches use a shared encoder and design task-specific decoders, which enable information exchange and facilitate task-specific feature learning during decoding, resulting in promising multi-task dense predictions.
However, the decoder-focused MTL models exhibit two primary limitations. First, the task-generic and task-specific patterns are entangled, thus models struggle to extract truly discriminative representations.
Second, the task-specific decoders are confined to activating single-task parameters, preventing the dynamic allocation of parameters across different tasks.
These limitations have prevented previous models from learning effective task-specific features and predictions.

To mitigate these challenges, the MoE architectures are leveraged in MTL. 
The MoE paradigm has shown considerable promise in addressing these inefficiencies by selectively activating only the most relevant portions of the model for a given task or input, thereby promoting computational efficiency while maintaining high capacity for diverse tasks. Despite this, existing MoE-based methods typically rely on a fixed expert allocation strategy across all tasks, which often leads to suboptimal performance. Specifically, the use of a static expert pool does not adequately accommodate the varying degrees of complexity across different tasks, potentially resulting in overfitting for simpler tasks or underfitting for more challenging ones. Furthermore, the rigid structure of expert selection fails to exploit task-specific features that could dramatically improve task performance for dense prediction.
These limitations highlight the need for a flexible MTL framework capable of fine-grained adaptation across tasks while minimizing trainable parameters.

In order to tackle the above-mentioned issues, we propose a novel multi-task mixture of fine-grained experts framework, named "FGMoE", which introduces adaptive expert allocation during the fine-tuning stage. FGMoE leverages a dynamic mixture of fine-grained experts, which are activated on a task-specific basis. Unlike conventional MoE methods that rely on a fixed expert pool, our framework selectively activates a subset of experts tailored to each task and region of interest in the image, allowing the model to capture both low-level and high-level spatial features with enhanced task-specific feature learning.
Specifically, our architecture contains three main designs: (1) \textbf{Intra-task experts}: we partition experts into finer-grained experts by splitting the intermediate hidden dimensions of MLP\cite{gMLP_2021}. Accordingly, while keeping the parameters unchanged, this design can activate more fine-grained intra-task experts for a more flexible and adaptable combination of activated experts.
Intra-task expert allows for a finer decomposition of different intra-task information and more precise mapping to different experts, while each expert can maintain a higher level of specialization.
(2) \textbf{Shared experts}:
We isolate some experts as always-active shared experts designed to capture and consolidate common information across different contexts of the same task.
By compressing common information into these shared experts, redundancy among other routing experts can be reduced. This improves parameter efficiency and ensures that each routing expert stays focused by concentrating on their unique aspects.
The increased flexibility of the intra-task experts and shared experts combinations also contributes to more accurate and targeted knowledge acquisition.
(3) \textbf{Global experts}: we introduce a novel global MoE layer designed specifically to aggregate and distribute task-relevant information across all dense prediction tasks. This global expert module serves as a cross-task knowledge repository, allowing beneficial information transfer while preventing harmful interference. Unlike traditional task-specific heads, our global MoE adaptively routes information based on both input characteristics and task requirements, enabling more flexible knowledge sharing. This design ensures that each task benefits from relevant information in other tasks while remaining protected from potentially harmful interference.

In summary, our contributions are as follows.
\begin{itemize}
  \item We propose a novel fine-tuning approach that adapts fine-grained experts and isolates some experts as shared ones for multi-task dense prediction tasks. By strategically allocating and activating fine-grained experts across spatial and channel dimensions, our method achieves task-specific specialization while preserving cross-task knowledge sharing.
  \item We introduce a global MoE coordination mechanism that dynamically balances information flow between task-specific and shared representations. This mechanism addresses a fundamental limitation in current MTL by adaptively modulating feature importance based on learned task compatibility matrices, reducing negative transfer while amplifying positive knowledge transfer across complementary tasks.
  \item Extensive experiments are conducted on two challenging multitask dense prediction datasets, $i.e.,$ NYUD-v2 and PASCAL-Context. The extensive results show that our method achieves competitive performance and only requires training small parameters ($i.e.,$ using fine-tuning) of the whole network.
\end{itemize}

\section{Related Work}
\label{sec:relate}

\subsection{Multi-Task Learning for Dense Prediction}
Multi-task learning (MTL) for dense prediction involves simultaneously learning multiple pixel-level prediction tasks, such as semantic segmentation, depth estimation, and surface normal prediction, within a single model \cite{MTL_survey_2021, mqtrans_xy}. MTL aims to leverage the inherent correlations between tasks to improve performance, efficiency, and generalization compared to training individual models for each task. However, MTL for dense prediction faces several challenges, including managing potential negative interference between tasks and balancing the learning process across tasks with varying characteristics. Recent research has explored various architectural and methodological approaches to address these challenges. 

Adapter Networks\cite{bhattacharjee2023vision, liang2022effective, xin2024vmt, jiang2024task} offer a parameter-efficient way to adapt large pre-trained models for multiple tasks by inserting small, trainable modules into the network layers while keeping the original weights frozen. Task-conditional approaches\cite{jiang2024task, huang2024going, xu2023multi} and prompting techniques\cite{taskprompter2023, lu2024prompt} explicitly tailor the network's processing based on the specific task, often using task prompts to guide the model. For example, Prompt Guided Transformer (PGT)\cite{lu2024prompt} incorporates task-specific prompts within the self-attention mechanism. Emerging Diffusion model-based methods, such as TaskDiffusion\cite{yang2025multitask} and DiffusionMTL\cite{ye2024diffusionmtl}, leverage the conditional diffusion process for joint denoising of multi-task labels, capturing task relations in the process. CCR\cite{yang2023contrastive} introduces feature-wise contrastive consistency to model cross-task interactions. Other works focus on improving MTL efficiency using various techniques. For example, Bi-MTPD\cite{shang2024efficient} proposes a framework that constructs a multi-task dense predictor using binarized neural networks, leading to significant acceleration and reduced memory footprint. 

\subsection{Mixture-of-Experts}
Mixture of Experts (MoE) have gained considerable attention for their ability to balance model capacity and efficiency through sparse activation of specialized sub-networks\cite{jacobs1993learning, jacobs1991adaptive, lrmoe2024, moe-tang2024}. The general principle behind MoE is to partition the network into multiple “experts,” each responsible for modeling a distinct subspace of the overall feature distribution. During inference or training, a lightweight "router" selects which experts to activate. Although MoE architectures have often been explored in natural language modeling\cite{du2022glam, fedus2022switch, dai2024deepseekmoe}, they have recently shown promise in multi-task dense prediction scenarios as well \cite{chen2023mod, liang2022mvit, taskexpert23, yang2024multi, jiang2024task}. TaskExpert\cite{taskexpert23} is the first work that introduces MoE into the decoder of the MTL model. MLoRE\cite{yang2024multi} improves the MoE in decoder using low-rank epxerts inspired by LoRA techniques, thus it attains higher scalability with less parameter overhead. 
Although not a pure MoE approach, Task-conditional adapter\cite{jiang2024task} adopts a parallel adapter structure that conditions task prompts for different dense tasks, similar in concept to MoE gating. 

\begin{figure*}[ht!]
\centering
  \includegraphics[width=0.95\textwidth]{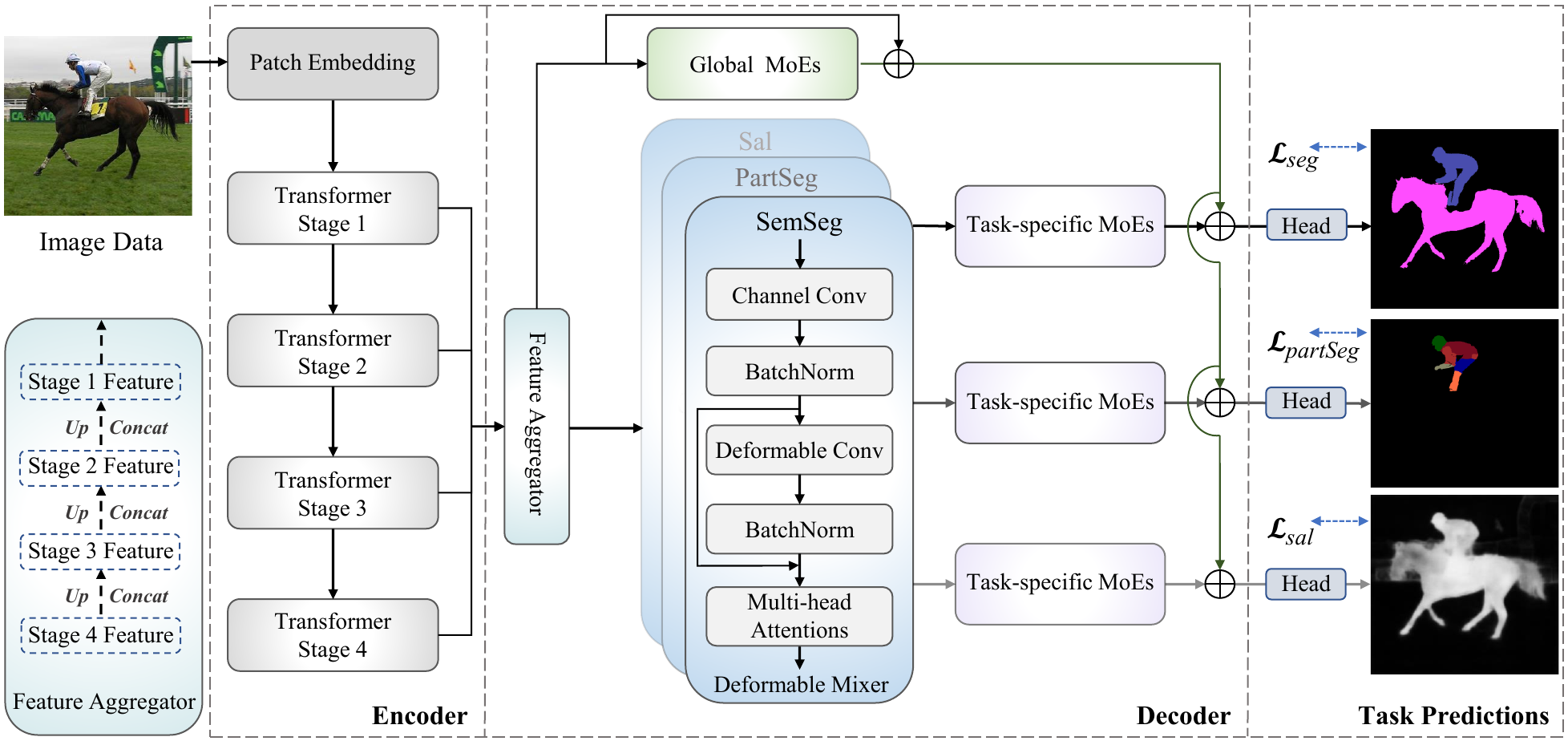}
  \caption{Overview of our FGMoE model. Given an input image, the Transformer Encoder extracts features that are initially aggregated by a Feature Aggregator. These features are then processed by both Global MoEs block and distinct task-specific Deformable Mixers, each paired with corresponding Task-specific MoEs. The outputs from these Task-specific MoEs are combined with the shared features from the Global MoE to produce the final task predictions. Within the Feature Aggregator (bottom-left), \textit{Up} and \textit{Concat} denote up-sampling operation and channel-wise concatenation. Symbol $\oplus$ denotes tensor addition. }\label{fig:over_view}
  \Description{fist figure.}
\end{figure*}

As the mentions above, we also leverage MoE for multi-task learning. Unlike TaskExpert\cite{taskexpert23} and MLoRE\cite{yang2024multi}, our FGMoE is restricted to fine-tuning specialist experts while keeping the backbone frozen, miminzing trainable parameters yet enabling task-level flexibility. Besides, FGMoE achieves more granular control over feature using the novel sparse expert activations across spatial and channel dimensions. Beyond Task-conditional Adapter\cite{jiang2024task}, we incorporate a global MoE that dynamically routes relevant features per pixel and task, allowing precise cross-task knowledge sharing.

\section{Approach}
\label{sec:method}

In this section, we first present an overview the proposed network architecture. Next, we detail the Decoder modules, which integrate the Feature Aggregator, Deformable Mixer, Global MoEs, and task-specific MoEs.
Finally, we briefly introduce the MTL training loss.

\subsection{Overall Architecture}
Th overall framework of FGMoE is illustrated in Figure~\ref{fig:over_view}. FGMoE follows an encoder-decoder paradigm comprising two main components: (1) Encoder: A shared Transformer-based backbone (i.e., Swin Transformer~\cite{swin}) extracts generic visual features from each input image, serving all tasks.
(2) Decoder: The decoder is composed of the Feature Aggregator, Deformable Mixer, Global MoEs, and Task-specific MoEs blocks. These blocks transform the encoder-extracted features into task-specific predictions.

\subsection{Feature Aggregator}
The feature aggregator integrates multi-scale features to create a shared feature map for all tasks, enabling the model to leverage information across scales and thereby improve MTL performance. 
As shown in Figure~\ref{fig:over_view}, the input image $X_{img} \in \mathbb{R}^{H \times W \times 3}$ ($H$, $W$, and $3$ denoting the height, width, and channels, respectively) is processed by the backbone ($i.e.,$ Swin Transformer) to generate four stages of image feature maps. Specifically, Stage 1 through Stage 4 produce image feature maps $X_{s1} \in \mathbb{R}^{\frac{H}{4} \times \frac{W}{4} \times C'}$, $X_{s2} \in \mathbb{R}^{\frac{H}{8} \times \frac{W}{8} \times 2C'}$, $X_{s3} \in \mathbb{R}^{\frac{H}{16} \times \frac{W}{16} \times 4C'}$, and $X_{s4} \in \mathbb{R}^{\frac{H}{32} \times \frac{W}{32} \times 8C'}$, respectively. These features are then up-sampled to the same resolution and concatenated along the channel dimension, yielding a feature map $X \in \mathbb{R}^{\frac{H}{4} \times \frac{W}{4} \times C''}$, where $C''=C'+2C'+4C'+8C'$.
We apply a linear projection to reduce the channel dimensionality (\textit{i.e., from C'' to C}).

\begin{figure*}[!ht]
\centering
  \includegraphics[width=0.790\textwidth]{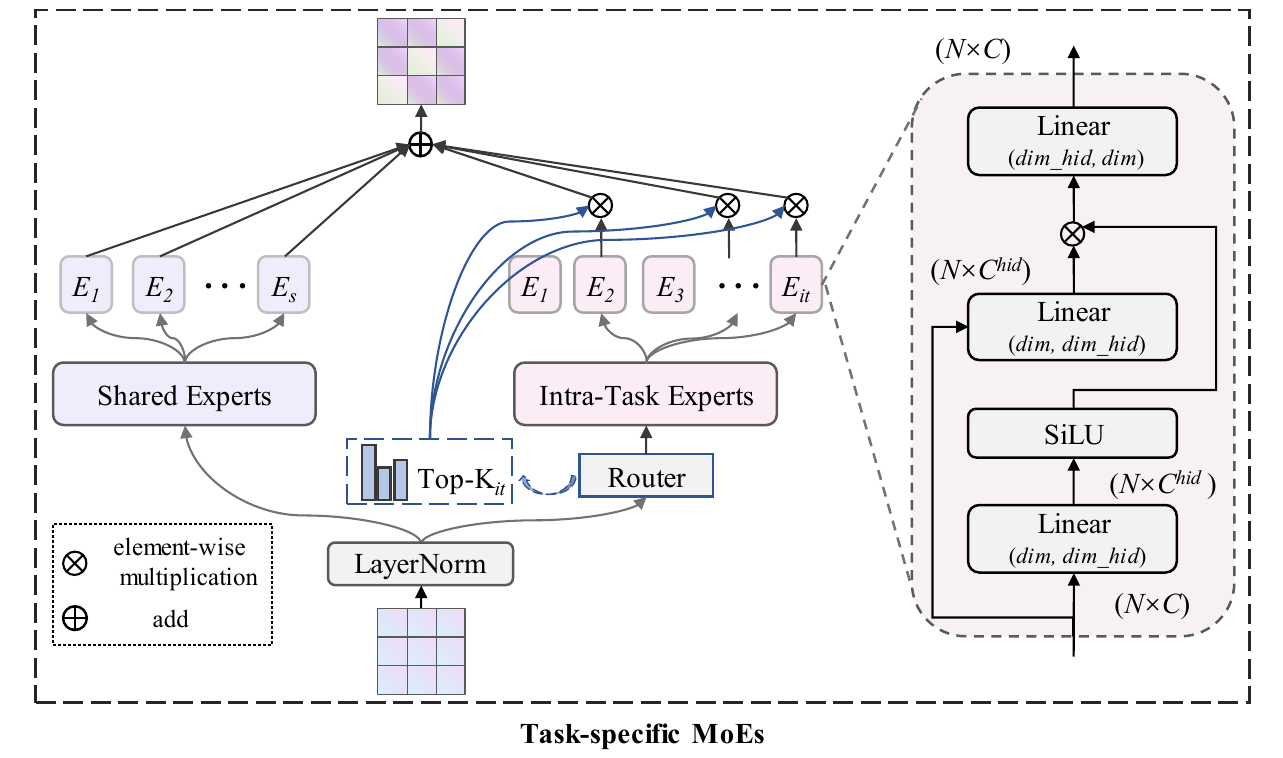}
  \caption{An overview of the task-specific MoEs block. Task-specific MoEs consists of the shared experts, Top-k, intra-task experts, and router. The core block of the task-specific MoEs block are (a) Shared Experts that reduce the knowledge redundancy among routing experts and (b) Intra-Task Experts that dynamically activate experts to select the task-aware features. For the composition structure of the expert block, please refer to the right side of the figure.}
  \label{fig:moe}
  \Description{2nd figure.}
\end{figure*}

\subsection{Deformable Mixer}
\noindent
\textbf{Channel Conv.}
The channel convolution employs a standard point-wise convolution (1$\times$1 kernel) to facilitate inter-channel communication.
Given the feature map $X\in \mathbb{R}^{\frac{H}{4} \times \frac{W}{4} \times C}$ obtained from the feature aggregator, the channel convolution fuses the information across channels:
\begin{equation}
X = W_p \cdot{X} + b,
\end{equation}
where $W_p$ is the point-wise convolution weight, and $b$ is the bias. 
We add a GELU activation ($\sigma(\cdot)$) and BatchNorm (BN) as follow:
\begin{equation}\label{eq:bn01}
 X =  \operatorname{BN}(\sigma(X)).
\end{equation}

\noindent
\textbf{Deformable Conv.}
To generate the relative offsets with respect to the reference point, the relative offsets are learned from the feature $X\in \mathbb{R}^{\frac{H}{4} \times \frac{W}{4} \times C}$ of upper layer.
$X_{i,j}$ is fed into the convolution operator to learn the corresponding offsets $\Delta_{(i,j)}$ for all reference points.
For each location point $(i,j)$ on $X_{i,j}$, the spatial deformable is mathematically represented as follows:
\begin{equation}
D_C(X_{i,j}) = \sum_{C'=0}^{C'-1}  W_d \cdot {X}((i,j)+\Delta_{(i,j)}),
\end{equation}
where the $W_d$ is a deformable weight, and the $\Delta_{(i,j)}$ is the learnable offset. 
A GELU activation, BatchNorm, and residual connection follow the deformable convolution:
\begin{equation}\label{eq:encoder}
  X_{C^\prime} = X_{i,j} + \operatorname{BN}(\sigma(D_C(X_{i,j}))),
\end{equation}
where $\sigma(\cdot)$ represents the non-linearity function (\textit{i,e.,} GELU).
Before LayerNorm (LN), we flatten the feature $X_{C'}\in \mathbb{R}^{\frac{H}{4} \times \frac{W}{4} \times C'}$ to a sequence with length ${N\times C'}$ ($N=\frac{H}{4} \times \frac{W}{4} $):
\begin{equation}\label{eq:reshape}
 X_{d'} =  \operatorname{LN}(\text{Flatten}(X_{C'})),
\end{equation}
where $X_{d'} \in \mathbb{R}^{N \times C'}$.

Then, the projected feature maps are passed through a transformer layer.
The multi-head self-attention (MHSA) is computed using the scaled-dot product:
\begin{align}\label{eq:mhsa}
    &\text{MHSA}(Q=X_{d'}, K=X_{d'}, V=X_{d'}) = \text{softmax}(\frac{QK^\top}{\sqrt{d_k}})V,
\end{align}
where $Q,K,V \in \mathbb{R}^{N \times C'}$ are the query, key and value matrices, and the output $\text{MHSA}(Q, K, V) = X_a \in \mathbb{R}^{N \times C'}$.

\subsection{Task-specific MoEs}
\textbf{Revisiting MoE.}
We first introduce the basic form of MoE architecture, like GShard\cite{GShared21}. Formally, we suppose that MoE contains $N$ experts and $M$ router networks, denoted as $E=\{E_1, E_2, \cdots,E_N\}$ and  $G=\{G_1, G_2, \cdots,G_M\}$, respectively.
A learnable gated network ($\textit{G}$) decides which part of the input to send to which experts ($E$):
\begin{equation}\label{eq:moe}
y =  \sum_{i=1}^N G_i(x)E_i(x).
\end{equation}
In Eq.~\ref{eq:moe}, all experts process each input and produce a weighted combination of their outputs, determined by the gating network. 
The gating network, typically implemented as a softmax-activated layer, dynamically routes inputs to the most relevant experts based on learned patterns. 
This mechanism is shown below.
\begin{equation}\label{eq:gating}
G_i(x) = \text{Softmax}(x \cdot W_g),
\end{equation}
where $W_g$ denotes the weight of the gated function.

\noindent
\textbf{Intra-Task Experts.}
As shown in Figure~\ref{fig:moe}, our task-specific MoE block comprises four key components: shared experts, Top-k gating, intra-task experts, and a router. 
The architecture isolates a subset of experts as shared (fine-grained experts) while adapting others for intra-task specialization. 
This design balances specialization and generalization in multi-task learning: intra-task experts capture task-specific nuances, while shared experts enable cross-task knowledge transfer. 
The model enhances expert specialization by integrating fine-grained experts (both shared and task-specific), ensuring tailored processing for distinct tasks without sacrificing shared feature learning.

We use shared experts and intra-task experts within finer-grained experts to acquire non-overlapping and focused task information.
Firstly, we apply a LayerNorm (LN) on the input feature:
\begin{equation}\label{eq:layernorm2}
 x_{a} =  \text{LN}(X_{a}).
\end{equation}
Let $x_a$ denotes the MoE block input. Formally, the output of the MoE layer is calculated as:

\begin{equation}\label{eq:ourmoe1}
y =   \sum_{i=1}^{N_s} E_i(x_{a}) + \sum_{i=1}^{N_{it}} G_i(x_{a})E_i(x_{a}),
\end{equation}

\begin{equation}\label{eq:ourmoe12}
G_{i} =  \frac{g_{i}}{\sum _{j=1} ^{N_{it}} g_{j}},
\end{equation}

\begin{equation}\label{eq:ourgating}
 g_i=\left\{
\begin{aligned}
s_i&, \quad s_i \in \text{Top-k}(\{s_{j}| 1 \le j \le N_{it} \}, K_{it}) \\
0&, \quad otherwise,
\end{aligned}
\right.
\end{equation}

\begin{equation}\label{eq:ourmoe3}
s_{i} =  \text{Sigmoid}(x_{a}^\top e_i),
\end{equation}
where $N_s$ denotes the number of shared experts.
$N_{it}$ denotes the number of intra-task experts.
$g_i$ denotes the gate value for the $i$-th expert.
$s_i$ is the intra-task feature to expert affinity. 
$e_i$ is the centroid of the $i$-th expert.
Top-k($s_i$,\textit{K}) represents the set of $K$ highest affinity scores calculated for intra-task feature and routing experts.
The intra-task expert design increases the number of non-zero gates to $N_{it}$, enabling full activation of all task-specific experts. 
This approach enhances combination flexibility by allowing dynamic aggregation of diverse expert outputs, where each input leverages a broader spectrum of specialized intra-task knowledge.

\noindent
\textbf{Shared Experts.}
Traditional routing strategies often assign features to different experts, which may require some common information. 
As a result, multiple experts will capture the shared task information in their respective parameters, leading to parameter redundancy. 
However, parameter redundancy among other routing experts can be mitigated if there are shared experts who specialize in capturing and integrating common knowledge in different contexts. 
This reduction in expert redundancy and parameter redundancy will help to build a more parameter-efficient model that contains more specialized experts.

To achieve this purpose, in addition to intra-task experts, we further group out $K_s$ experts as shared experts. 
In practice, we fix the number of shared experts to 2.
These shared experts do not use routing computation, so each feature will be deterministically assigned to these shared experts. 
After integrating the shared expert isolation policy, the formula for shared experts is as follows:
\begin{equation}\label{eq:sharedmoe}
y_s =   \sum_{i=1}^{N_s} E_i(x_{a}).
\end{equation}

\subsection{Global MoEs}
The global MoEs block uses the same MoEs architecture as the Task-specific MoEs block.
With the fine-grained task-specific experts, the output of a Global MoEs block can be expressed as:
\begin{equation}\label{eq:globalmoe}
y_g = \text{Global-MoEs}(X) + X,
\end{equation}
where the feature map $X$ is derived from the output of the feature aggregator, as depicted in Figure~\ref{fig:over_view}.
The $y_g$ is then added to the outputs of multiple task-specific MoEs respectively, and the result of the addition is fed into the corresponding head.

\subsection{MTL Training Objective}
Training our FGMoE framework involves optimizing a multi-objective function that balances task-specific losses, encourages effective expert specialization, and promotes beneficial knowledge transfer.
Here forms a total loss for multi-task:
\begin{equation}\label{equ:loss}
	\begin{split}
	    \begin{aligned}
    {\mathcal L_{total}} &= \sum_{t=1}^{T} \beta_{t}{\mathcal L_{t}}, \\
       \end{aligned}
    \end{split}
\end{equation}
where $\beta_t$ and $\mathcal L_{t}$ denotes the set of hyper-parameters ($i.e.,$ balancing factors) and task-specific losses in \cref{equ:lossdetal}.
$T$ denotes the total number of tasks ($t \in [1,T]$).
\begin{equation}\label{equ:lossdetal}
  \begin{split}
  \begin{aligned}
    {\mathcal L_{t}} & \in \{\mathcal L_{seg}, \mathcal L_{partseg}, \mathcal L_{sal}, \mathcal L_{depth}, \mathcal L_{normal}, {\mathcal L_{bound}}\},\\
       {\beta_{t}} & \in \{\beta_{seg}, \beta_{partseg}, \beta_{sal}, \beta_{depth}, \beta_{normal}, {\beta_{bound}}\}.
    \end{aligned}
    \end{split}
\end{equation}

\section{Experiments}
\label{sec:exp}
To evaluate the proposed method, we conduct extensive experiments on two dense prediction datasets, $i.e.,$ NYUD-v2\cite{NYUD2012} and PASCAL-Context\cite{pascal2014}. 
The HRNet18~\cite{HRnet_19}, ViT-Large (ViT-L)~\cite{ViT2021}, Swin-Tiny (Swin-T)~\cite{swin}, and Swin-Large (Swin-L)\cite{swin} are used as backbones.
These datasets provide diverse scene characteristics and varying task complexities, enabling a thorough assessment of our approach's performance and parameter efficiency.

\begin{table*}[!t]
\centering
\caption{Experimental results on PASCAL-Context dataset. ‘$\downarrow$’: lower is better. ‘$\uparrow$’: Higher is better. $\Delta_m$ denotes the average per-task performance drop (the higher, the better). $\dagger$: Evaluated with the official code.}
\label{tab:sota_comparison_pascal}
\begin{tabular}{lcccccccccc}
\toprule[0.1em]
 \multirow{2}*{Model}  & \multirow{2}*{Backbone}  & {SemSeg}   &PartSeg &Sal  & Normal & Bound &\multirow{2}*{$\Delta_m[\%]$$\uparrow$}  &FLOPs &Params\\
     &  & (mIoU)$\uparrow$  & (mIoU)$\uparrow$  &(maxF)$\uparrow$  &(mErr)$\downarrow$  &(odsF)$\uparrow$ & &(G)$\downarrow$ &(M)$\downarrow$\\
\hline
Single task baseline           &HRNet18 &62.23 &61.66 &85.08 &13.69 &73.06 &0.00 &- &- \\
MTI-Net~\cite{Mti-net_2020}    &HRNet18 &61.70 &60.18 &84.78 &14.23 &70.80 &-2.10  &161 &128\\
ATRC~\cite{atrc_2021}          &HRNet18 &57.89 &57.33 &83.77 &13.99 &69.74 &-4.45 &216 &96\\
DeMT\cite{xu2023demt}          &HRNet18 &59.23 &57.93 &83.93 &14.02 &69.80 &-3.79  &- &-\\
FGMoE (Ours)                   &HRNet18 &60.52 &59.01 &84.81 &13.51 &72.10 &-1.47  &204 &91\\
\hdashline
Single task baseline        &ViT-L &81.62 &72.21 &84.34 &13.59 &76.79 &0.00    &- &- \\
PAD-Net~\cite{Pad-net_2018} &ViT-L &78.01 &67.12 &79.21 &14.37 &72.60 &-5.72 &773  &330\\
MTI-Net~\cite{Mti-net_2020} &ViT-L &78.31 &67.40 &84.75 &14.67 &73.00 &-4.62 &774  &851\\
ATRC~\cite{atrc_2021}       &ViT-L &77.11 &66.84 &81.20 &14.23 &72.10 &-5.50 &871  &340\\
InvPT~\cite{InvPT_2022}     &ViT-L &79.03 &67.61 &84.81 &14.15 &73.00 &-3.61 &669  &423\\
TaskPrompter~\cite{taskprompter2023} &ViT-L &80.89 &68.89 &84.83 &13.72 &73.50 &-2.03 &497 &401\\
TaskExpert~\cite{taskexpert23}       &ViT-L &80.64 &69.42 &84.87 &13.56 &73.30 &-1.74 &622 &420\\
MLoRE~\cite{yang2024multi}            &ViT-L &81.41 &70.52 &84.90 &13.51 &75.42 &-0.62 &571 &407\\
SEM~\cite{huang2024going}             &ViT-L &81.66 &69.90 &84.95 &13.39 &73.80 &-0.98 &- &-\\
FGMoE(Ours)               &ViT-L &\textbf{82.19} &\textbf{71.32} &\textbf{85.15} &\textbf{13.25} &\textbf{75.94} &0.39 &503 &320\\
\hdashline
Single task baseline      &Swin-L &79.26 &68.92 &83.84 &14.28 &74.50 &0.00   &- &-\\
InvPT~\cite{InvPT_2022}   &Swin-L &78.53 &68.58 &83.71 &14.56 &73.60 &-0.95 &500$^\dagger$ &314$^\dagger$ \\
FGMoE (Ours)             &Swin-L &\textbf{78.55} &\textbf{68.61} &\textbf{83.94} &\textbf{14.05} &\textbf{74.30} &0.02 &425 &206 \\
\bottomrule[0.1em]
  \end{tabular}
\end{table*}

\subsection{Experimental Setup}
\textbf{Datasets.} NYUD-V2 consists of video sequences of various indoor visual scenes recorded by RGB and depth cameras. NYUD-V2 has 1449 densely labeled pairs of aligned RGB and depth images, with 464 new scenes taken from 3 cities, which are 795 training and 654 testing. NYUD-V2 is mainly used for semantic segmentation ('SemSeg'), depth estimation ('Depth'), surface normal estimation ('Normal'), and boundary detection ('Bound') tasks.
PASCAL-Context has a total of 459 annotations containing 10103 images, of which 4998 are used in the training set and 5105 in the validation set.
PASCAL-Context is labeled for semantic segmentation, human parts segmentation ('PartSeg'), saliency estimation ('Sal'), surface normal estimation ('Normal'), and boundary detection.

\noindent
\textbf{Evaluation metric.}
We adopt four evaluation metrics to evaluate the performance of the multi-task models.
(1) The mean Intersection over Union (mIoU) is used to evaluate the semantic segmentation and human part segmentation tasks.
(2) The root mean square error (rmse) is used to evaluate the depth estimation task.
(3) The mean Error (mErr) is used to evaluate the surface normals task.
(4) The optimal dataset scale F-measure (odsF) is used to evaluate the edge detection task.
(5) The maximum F-measure (maxF) is used to evaluate the saliency estimation task.
(6) For multi-task learning, it’s difficult to measure the performance of a model with a single metric; we introduce the average per-task performance drop ($\Delta_m$) metric.
The average per-task performance drop ($\Delta_m$) is used to quantify multi-task performance. $\Delta_m=\frac{1}{T} \sum_{i=1}^T(F_{m,i}-F_{s,i})/F_{s,i}\times100\%$, where $m$, $s$ and $T$ mean multi-task model, single task baseline and task numbers.
$\Delta_m$: the higher, the better.

\noindent
\textbf{Training Hyper-Parameters}.
The hyper-parameters of Eq.~\ref{equ:lossdetal}, their values are
$\beta_{seg}$ =1.0, $\beta_{depth}$ =1.0, $\beta_{normal}$ = 10.0, $\beta_{bound}$ =50.0, $\beta_{partseg}$ =2.0, and $\beta_{sal}$ =5.0.
Each MoE layer consists of 2 shared expert and 6 routed experts, where the intermediate hidden dimension of each expert is 256.
We employ the SGD optimizer with hyper-parameters set to learning rate $0.001$, and weight decay $0.0005$.

\subsection{Comparison with State-of-the-Art}
\noindent\textbf{Full-model training.} We implement FGMoE with HRNet18, ViT-L, and Swin-L backbones and compare it against to recent state-of-the-art MTL methods on the PASCAL-Context and NYUDv2 datasets (see Tables~\ref{tab:sota_comparison_pascal} and \ref{tab:sota_comparison_NYUD}). For a more comprehensive comparison on PASCAL-Context dataset, we re-implement several leading CNN-based multi-task learning methods on the ViT-L backbone.

Table~\ref{tab:sota_comparison_pascal} presents a comprehensive comparison of our proposed FGMoE method against current state-of-the-art approaches on the PASCAL-Context dataset using HRNet18, ViT-L, and Swin-L backbones. When employing HRNet18, FGMoE achieves the highest mIoU for semantic segmentation (68.12) and parts segmentation (63.03), while also maintaining competitive performance in saliency detection (83.82), normal estimation (14.17), and boundary detection (72.12). This level of performance is accomplished with fewer FLOPs (204G) and parameters (91M) compared to many existing methods. Notably, upon switching to the ViT-L backbone, FGMoE attains new state-of-the-art results across all tasks, surpassing strong baselines such as MLoRE and TaskExpert. These gains are accompanied by lower computational overhead (503G FLOPs) and reduced model size (320M parameters), indicating that FGMoE is capable of fully leveraging the representational capacity of large-scale transformers while remaining resource-efficient. Furthermore, when using the Swin-L backbone, FGMoE continues to outperform InvPT~\cite{InvPT_2022} in both predictive accuracy and computational metrics, achieving improved scores on all tasks with fewer FLOPs (425G vs. 500G) and parameters (206M vs. 314M). This consistent superiority under multiple architectural settings illustrates FGMoE’s adaptability and effectiveness in tackling multi-task vision challenges.

\begin{table}[t]
\small
\caption{Experimental results on NYUD-v2 dataset with ViT-L backbone. '$\downarrow$': lower is better. '$\uparrow$': Higher is better. \label{tab:stoa_nyud_moe}}
\label{tab:sota_comparison_NYUD}
\centering
\begin{tabular}{lcccccc}
\toprule[0.1em]
 \multirow{2}*{Model}  &SemSeg &Depth & Normal & Bound &\multirow{2}*{$\Delta_m[\%]$}\\
       &(mIoU)$\uparrow$   &(rmse)$\downarrow$   & (mErr)$\downarrow$  &(odsF)$\uparrow$\\
\hline
Single task                 &56.77 &0.5141 &18.56 &78.93 &0.00\\
InvPT\cite{InvPT_2022}              &53.56 &0.5183 &19.04 &78.10 &-2.52\\
TaskPrompter\cite{taskprompter2023} &55.30 &0.5152 &18.47 &78.20 &-0.81\\
TaskExpert\cite{taskexpert23}       &55.35 &0.5157 &18.54 &78.40 &-0.84\\
MLoRE\cite{yang2024multi}            &55.96 &0.5076 &18.33 &78.43 &0.11\\
FGMoE (Ours)                         &56.16 &0.5071 &19.19 &78.70 &-0.58\\
\bottomrule[0.1em]
  \end{tabular}
\end{table}

\begin{table*}[ht]
\centering
\caption{Fine-tuning results on PASCAL-Context. “Single Inference For All Tasks” indicates whether the model allows all tasks to be performed simultaneously. $\Delta_m$ denotes the average per-task performance drop. ‘$\downarrow$’: lower is better. ‘$\uparrow$’: Higher is better. }
\label{tab:finetune:pascal}
\begin{tabular}{lccccccc}
\toprule[0.1em]
 \multirow{2}*{Model}   & {SemSeg}   &PartSeg &Sal  & Normal  &\multirow{2}*{$\Delta_m[\%]$$\uparrow$}  &Trainable  &Single Inference\\
     & (mIoU)$\uparrow$  & (mIoU)$\uparrow$  &(mIoU)$\uparrow$  &(mErr)$\downarrow$& &Parameters (M) &For All Tasks\\
\noalign{\smallskip}
\hline
\noalign{\smallskip}
Single task baseline          &67.21 &61.93 &62.35 &17.97 &0.00 &112.62    &\ding{55}\\
MTL-Tuning Decoder Only       &65.09 &53.48 &57.46 &20.69 &-9.95 &1.94  &\ding{51}\\
MTL-Full Fine Tuning          &67.56 &60.24 &65.21 &16.64 &+2.23 &30.06 &\ding{51}\\
\hline
Adapte~\cite{he2022adapter}     &69.21 &57.38 &61.28 &18.83 &-2.71 &11.24 &\ding{55}\\
Bitfit~\cite{bitfit2022}        &68.57 &55.99 &60.64 &19.42 &-4.60 &2.85 &\ding{55}\\
VPT-shallow~\cite{VPTshallow2022}&62.96 &52.27 &58.31 &20.90 &-11.18 &2.57 &\ding{55}\\
VPT-deep~\cite{VPTshallow2022} &64.35 &52.54 &58.15 &21.07 &-10.85 &3.43 &\ding{55}\\
Compactor~\cite{compacter2021} &68.08 &56.41 &60.08 &19.22 &-4.55 &2.78 &\ding{55}\\
Compactor++~\cite{compacter2021} &67.26 &55.69 &59.47 &19.54 &-5.84 &2.66 &\ding{55}\\
LoRA~\cite{hu2022lora}         &70.12 &57.73 &61.90 &18.96 &-2.17 &2.87 &\ding{55}\\
VL-Adapter~\cite{VLAdapter}    &70.21 &59.15 &62.29 &19.26 &-1.83 &4.74 &\ding{55}\\
HyperFormer~\cite{hyperFormer2021} &71.43 &60.73 &65.54 &17.77 &+2.64 &72.77 &\ding{55}\\
Polyhistor~\cite{Polyhistor22}  &70.87 &59.54 &65.47 &17.47 &+2.34 &8.96 &\ding{55}\\
MTLoRA~\cite{MTLoRA24} (r = 16) &68.19 &58.99 &64.48 &17.03 &+1.35 &4.95 &\ding{51}\\
FGMoE+Swin-T (Ours) &67.37 &56.87 &64.12 &15.67 &+1.96 &3.50  &\ding{51}\\
\hdashline
DITASK+Swin-L   &76.23 &\textbf{67.53} &64.07 &16.90 &+7.79 &7.13 &\ding{51}\\
FGMoE+Swin-L (Ours) &\textbf{77.34} &64.16 &\textbf{64.13} &\textbf{16.40} &\textbf{+7.89} &4.70  &\ding{51}\\
\bottomrule[0.1em]
  \end{tabular}
\end{table*}

Table~\ref{tab:sota_comparison_NYUD} reports performance on the NYUD-v2 using the ViT-L backbone. Similar to the observations on PASCAL-Context, FGMoE surpasses existing methods in semantic segmentation (mIoU=56.16), depth (rmse=0.5071), and boundary (odsF=78.70). The strong performance across both PASCAL-Context and NYUD-v2 underscores the effectiveness and robustness of the proposed FGMoE, further highlighting its adaptability and competitiveness in MTL.

\noindent\textbf{Decoder-only Fine-Tuning.} In addition to training both the encoder and decoder, we also apply FGMoE in a decoder-only fine-tuning setting, which significantly reduces the number of trainable parameters compared to full model training. The results presented in Table~\ref{tab:finetune:pascal} summarizes the results of decoder-only fine-tuning on the PASCAL-Context dataset. As expected, basic decoder fine-tuning (MTL-Tuning Decoder Only) reduces the number of trainable parameters but leads to decreased performance across multiple tasks ($\Delta_m=-9.95\%$) compared to the single task baseline. Full model fine-tuning (MTL - Full Fine Tuning) recovers some performance but incurs significantly higher parameter costs. When comparing the performance of various adapter- and prompt-based methods, FGMoE consistently demonstrates strong results while maintaining a low parameter footprint. Specifically, FGMoE+Swin-L achieves the best average performance improvement ($\Delta_m=+8.01\%$), outperforming DITASK+Swin-L by 0.22\% in $\Delta_m$ while requiring fewer trainable parameters (4.7M vs. 7.13M). Notably, both FGMoE+Swin-T and FGMoE+Swin-L enable single inference for all tasks, highlighting the practicality of FGMoE for efficient multi-task deployment.

\begin{table}[ht]
  \caption{Decoder fine-tuning results on NYUD-v2 for Semantic Segmentation (SemSeg) and Depth estimation tasks. Swin-T is used as backbone. $\Delta_m$ denotes the average per-task performance drop. ‘$\downarrow$’: lower is better. ‘$\uparrow$’: Higher is better. }
  \label{tab:finetun_nyud}
  \centering
  \small
  \tabcolsep=0.14cm
\begin{tabular}{cccccc}
\toprule[0.1em]
 \multirow{2}*{Method}     &{SemSeg}   &Depth &$\Delta_m$ &{Trainable }\\
      &(mIoU)$\uparrow$   &(rmse)$\downarrow$ &(\%)$\uparrow$  &{Params (M)}\\
\noalign{\smallskip}
\hline
\noalign{\smallskip}
{Single task baseline}     &33.18   &0.667  &0.00   &112.62 \\
{MTL-Tuning Decoder Only}  &28.37   &0.832  &-19.61 &1.00\\
{MTL-Full Fine-Tuning}     &35.29   &0.734  &-1.84  &28.5 \\
\hline
MTLoRA~\cite{MTLoRA24}     &37.18   &0.635  &+8.42  &6.26 \\
\hline
Single-task DITASK         &44.01   &0.644  &+18.04 &1.61\\
DITASK~\cite{ditask2025}   &43.85   &\textbf{0.606}  &+20.65 &1.61 \\
\hline
FGMoE-Tuning (Ours)        &\textbf{44.71 } &0.618  &\textbf{+21.02 }&2.50 \\
\bottomrule[0.1em]
\end{tabular}
\end{table}

Table~\ref{tab:finetun_nyud} demonstrate that the proposed FGMoE-Tuning achieves the best balance between performance and parameter efficiency: it attains the highest average improvement of $+21.02\%$ with fewer than 2.5 million trainable parameters. Notably, FGMoE surpasses single-task DITASK and even outperforms DITASK~\cite{ditask2025} in semantic segmentation (44.71 vs. 43.85 mIoU), highlighting both its efficacy in preserving high-level semantic features and its efficiency in parameter usage. This strong performance underscores FGMoE’s capacity to deliver state-of-the-art results under stringent parameter constraints, making it well-suited for practical multi-task settings.

\subsection{Ablation Studies}
\begin{table}[ht!]
  \caption{Ablation study with Swin-T backbone on NYUD-v2 dataset.
  Deformable mixer (DM), global MoEs (GM), and task-specific MoEs (TSM) are the components of our method. The {w/} indicates {"with"}.}
  \label{tab:ablation_component}
  \centering
  \small
\begin{tabular}{cccccc}
\toprule[0.1em]
 \multirow{2}*{Components}     &{SemSeg}   &Depth &Normal  &Bound  &$\Delta_m$\\
  &(mIoU)$\uparrow$ &(rmse)$\downarrow$ &(mErr)$\downarrow$ &(odsF)$\uparrow$ &(\%)$\uparrow$\\
\noalign{\smallskip}
\hline
\noalign{\smallskip}
Single task baseline   &42.92	&0.6104	&20.94	&76.2  &0.00\\
Multi-task baseline    &38.78   &0.6312 &21.05  &75.6   &-3.74\\
\textit{w/} DM+GM      &45.02	&0.5741 &20.65  &76.7   &3.19\\
\textit{w/} DM+GM+TSM  &\textbf{47.12} &\textbf{0.5616} &\textbf{20.17} &\textbf{77.1} &5.17\\
\bottomrule[0.1em]
\end{tabular}
\end{table}

\textbf{Ablation on components.} We conduct the ablation studies using Swin-T backbone on NYUD-v2 dataset. As shown in Table~\ref{tab:ablation_component}, we investigate the individual contribution of each component in our FGMoE.
The results validate the effectiveness of our deformable mixer, global MoEs and Task-specific MoEs. 
The design of deformable mxier (DM) block is inspired by DeMT~\cite{xu2023demt}, and we refine its structure.
This analysis allowed us to gain insight into the importance and effectiveness of each component in the overall performance of the method. 
In Table~\ref{tab:ablation_component}, we observe that the 
Combining task-specific moe (TSM) yields additional performance gains in all metrics. The full model achieves an optimal semantic segmentation mIoU of 47.12 (+8.34 improvement \textit{V.S.} multi-task baseline). The consistency improvements added by each component demonstrate their complementary properties in our architecture.
The evident performance improvement from the multi-task baseline to the full model validates our design choices and confirms that each component contributes meaningfully to the overall performance. Notably, the remarkable improvement in semantic segmentation indicates that our MoEs is particularly effective in capturing semantic texture.

\begin{table}[ht!]
  \caption{ Ablation study for expert numbers with Swin-T backbone on NYUD-v2 dataset. The number of shared experts in task-specific MoEs is fixed to be 2.}
  \label{tab:ablation_expernum}
  \centering
\begin{tabular}{cccccc}
\toprule[0.1em]
 \multirow{2}*{Experts} &{SemSeg}   &Depth &Normal  &Bound  &$\Delta_m$\\
   &(mIoU)$\uparrow$  &(rmse)$\downarrow$   &(mErr)$\downarrow$  &(odsF)$\uparrow$ &(\%)$\uparrow$\\
\noalign{\smallskip}
\hline
\noalign{\smallskip}
2    &45.91	   &0.5722  &21.01   &76.9  &3.41\\
4    &46.92	   &0.5645  &20.24   &77.1  & 4.24\\
6    &47.12	   &0.5616  &20.17   &77.1  &5.17\\
8    &\textbf{48.09}    &\textbf{0.5611}  &20.04   &\textbf{77.3} &6.35\\
16   &47.82    &0.5651  &\textbf{19.84}   &77.2  &6.15\\
\bottomrule[0.1em]
\end{tabular}
\end{table}

\noindent
\textbf{Ablation on expert numbers.}
To investigate the optimal expert configuration for our approach, we conduct a detailed ablation study on the number of experts in our task-specific Mixture of Experts (MoEs) while keeping the number of shared experts fixed at 2. Table~\ref{tab:ablation_expernum} presents the performance across four dense prediction tasks on the NYUD-v2 dataset using a Swin-T backbone.
Increasing the number of experts from 2 to 8 leads to an increase in performance for all metrics, but increasing from 8 to 16 results in a slight decrease for almost all metrics, except for mErr of normal prediction. This result shows that simply increasing the number of experts does not necessarily lead to performance improvement, and may even lead to inefficient routing or training difficulties.

\begin{table}[ht!]
  \caption{ Ablation on Top-$K$ on NYUD-v2. The numbers of shared experts and intra-task experts are fixed to be 2 and 6.}
  \label{tab:ablation_topk}
  \centering
\begin{tabular}{cccccc}
\toprule[0.1em]
 \multirow{2}*{Top-$K$}     &{SemSeg}   &Depth &Normal  &Bound  &$\Delta_m$\\
 &(mIoU)$\uparrow$  &(rmse)$\downarrow$  &(mErr)$\downarrow$  &(odsF)$\uparrow$ &(\%)$\uparrow$\\
\noalign{\smallskip}
\hline
\noalign{\smallskip}
2   &\textbf{47.29}	 &0.5742  &\textbf{20.01}  &77.0 &5.23 \\
3   &47.12	 &0.5616  &20.17  &77.1  &5.17\\
4   &47.21	 &0.5602  &20.10  &\textbf{77.2}  &5.60\\
6   &47.18	 &\textbf{0.5594}  &20.16  &77.0  &5.55\\
\bottomrule[0.1em]
\end{tabular}
\end{table}

\noindent
\textbf{Ablation on Top-$K$ numbers.}
For each task, the intra-task MoEs router learns gating values for these experts and activates the Top-$K$ experts according to the gating values.
In this ablation study, we analyze the impact of varying the Top-$K$ parameter on multi-task-intensive prediction in a mixed-expert (MoE) architecture. Table~\ref{tab:ablation_topk} shows the results on the NYUD-v2 dataset using a model containing 2 shared experts and 6 intra-task experts, along with the number of $K$ taken differently (\textit{i.e.,} $K \in \{2,3,4,6\}$) during training.
The Top-$K$ parameter is a critical design choice in MoEs, determining the sparsity of expert activation and consequently affecting both computational efficiency and representational capacity.
Table~\ref{tab:ablation_topk} shows that higher $K$ values do not carry better performance. This manifestation suggests that for tasks requiring precise spatial localization ($e.g.,$ segmentation and boundary detection), the focus specialization achieved by Top-3 routing may be beneficial, thus reducing potential conflicts between expert predictions.

\noindent
\subsection{Qualitative Results}
\begin{figure}[ht]
\centering
  \includegraphics[width=0.49\textwidth]{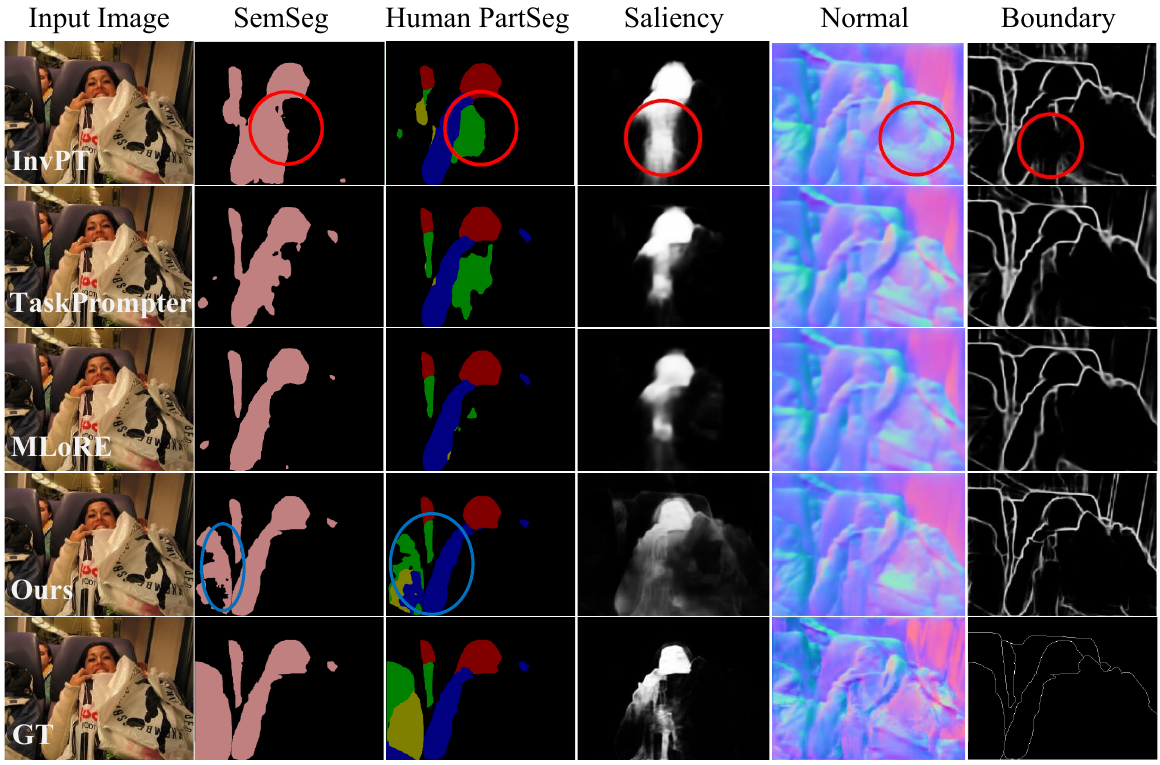}
  \caption{ Qualitative results on 5-task PASCAL-Context dataset comparing our approach with InvPT~\cite{InvPT_2022}, TaskPrompter~\cite{taskprompter2023}, and MLoRE~\cite{yang2024multi}. {Note that they are marked in blue and red circles.} The results illustrate the effectiveness of our model in capturing both semantic and fine-grained details. Best viewed in color and zoom.
  }
\label{fig:vis_diff_methods}
\end{figure}
Our extensive visual evaluation demonstrates the superior capability of our FGMoE model compared to several leading methods in multi-task visual understanding. In Figure \ref{fig:vis_diff_methods}, we present visualization results alongside those from InvPT~\cite{InvPT_2022}, TaskPrompter~\cite{taskprompter2023}, and MLoRE~\cite{yang2024multi} (all marked in blue and red circles), revealing FGMoE's notable advantages in both semantic comprehension and fine-grained detail preservation.
The qualitative results clearly demonstrate FGMoE's ability to extract and represent semantic information across diverse scenes more efficiently. While TaskPrompter~\cite{taskprompter2023} shows competence in semantic segmentation through its unified multi-task framework, it frequently maintain contextual coherence in complex scenes with multiple interacting objects.
A critical advancement of our FGMoE architecture is its exceptional ability to preserve fine-grained visual details while maintaining semantic consistency. The strength is particularly evident when comparing FGMoE to MLoRE~\cite{yang2024multi}: although MLoRE employs multiple mixture-of-experts modules, it often generates smoothed boundaries and loses important textural details. By contrast, FGMoE consistently cures both high-level semantic information and fine-grained details--a balance that previous methods have struggled to achieve.

\section{Conclusion}
\label{sec:conclusion}
In this work, we introduced a novel Fine-Grained MoE (FGMoE) architecture that addresses fundamental limitations in existing MoE frameworks for dense predictions. Our approach makes three key technical contributions: intra-task experts for fine-grained specialization, shared experts for common information consolidation, and global experts for cross-task knowledge transfer. 
In addition, with fine-tuned settings, our method achieves competitive performance with comparable parametric efficiency.
Extensive experiments demonstrate the superiority of our approach in achieving highly accurate multi-task dense predictions across multiple datasets ($i.e.,$ NYUD-v2 and PASCAL-Context).

Future work will explore promising directions for top-$k$ routing. Dynamic adaptation of the $k$ value based on input complexity and multi-task requirements may further optimize the balance between computational efficiency and representation capacity.

\bibliographystyle{ACM-Reference-Format}
\bibliography{ref-mtl}

\end{document}